\documentclass[review,onefignum,onetabnum]{article}

\usepackage[utf8]{inputenc} 
\usepackage[T1]{fontenc}    
\usepackage{hyperref}       
\usepackage{url}            
\usepackage{booktabs}       
\usepackage{amsfonts}       
\usepackage{nicefrac}       
\usepackage{microtype}      
\usepackage{subcaption}
\usepackage{wrapfig}
\usepackage[backend=biber,style=numeric]{biblatex}
\usepackage{theorem,ifthen}
\usepackage{algorithm}
\usepackage[noend]{algpseudocode}
\bibliography{biblio} 

\usepackage{lineno}
\usepackage{tikz}

\usepackage{amsmath,amsfonts,bm}

\def\eqref#1{equation~\ref{#1}}
\def\Eqref#1{Equation~\ref{#1}}

\def\1{\bm{1}}

\DeclareMathAlphabet{\mathsfit}{\encodingdefault}{\sfdefault}{m}{sl}
\SetMathAlphabet{\mathsfit}{bold}{\encodingdefault}{\sfdefault}{bx}{n}

\newcommand{\hf}{{\frac 12}}

\newcommand{\bfA}{{\bf A}}

\newcommand{\bfH}{{\bf H}}
\newcommand{\bfI}{{\bf I}}

\newcommand{\bfL}{{\bf L}}

\newcommand{\bfW}{{\bf W}}
\newcommand{\bfX}{{\bf X}}

\newcommand{\bfx}{{\bf x}}
\newcommand{\bfy}{{\bf y}}

\newcommand{\bfd}{{\bf d}}

\newcommand{\bfv}{{\bf v}}
\newcommand{\bfw}{{\bf w}}
\newcommand{\bfz}{{\bf z}}

\newcommand{\bfepsilon}{{\boldsymbol \epsilon}}
\newcommand{\bftheta}{{\boldsymbol \theta}}

\newcommand{\bfSigma}{{\boldsymbol \Sigma}}

\title{A-Optimal Active Learning}

\author{Tue Boesen \small{\textsuperscript{1}}\\
{\tt\small Tue.Boesen@protonmail.com}
\and
Eldad Haber \small{\textsuperscript{1}}\\
{\tt\small EldadHaber@gmail.com} \\
\small \textsuperscript{1} University of British Columbia, Vancouver, Canada
\\
}

\begin{document}

\maketitle

\begin{abstract}
In this work we discuss the problem of active learning. We present an approach that is based
on A-optimal experimental design of ill-posed problems and show how one can optimally label a data set by partially probing it,  and use it to train a deep network. We present two approaches that make different assumptions on the data set. The first is based on a Bayesian interpretation of the semi-supervised learning problem with the graph Laplacian that is used for the prior distribution and the second is based on a frequentist approach, that updates the estimation of the bias term based on the recovery of the labels. We demonstrate that this approach can be highly efficient for estimating labels and training a deep network.
\end{abstract}

\section{Introduction}	

In this paper we discuss the problem of active learning  using the framework of optimal experimental design.
In this setting, we assume  to have a  data set $\bfX = [\bfx_1,\ldots,\bfx_n]$ where each $\bfx_j$ is a vector in $\mathbb{R}^m$. We also assume that for each $\bfx_j$ there is an associated real number $y_j$ that we note as the label. 
The relation between $\bfx$ and $\bfy$ is, an unknown mapping, $F:\mathbb{R}^m\rightarrow \mathbb{R}$
\begin{eqnarray}
F(\bfx) = \bfy.
\end{eqnarray}
For supervised learning, we assume that we are given pairs $\{\bfx_i,\bfy_i\}$ that are then used for training. In the active learning context we assume that we are given the data $\bfX$ with either partial or no labels.
The active learning problem seeks to optimize which data points in a data set should be labeled in order to approximately reconstruct the function, $F(\bfx)$. To this end,  we assume that we are able to ask an "oracle" for the label of a particular data point, however, this comes at a cost, which may be significant \cite{bouneffouf:hal-01069802,ALreview2010}. One therefore aims to develop a model with a small set of data.

There are two possible ways to tackle the problem \cite{OneillActive}. The first assumes a particular learning model typically with a small Vapnik–Chervonenkis dimension, for example, Support Vector Machine or a shallow neural network. This model can be used in order to decide which data should be labeled next in order to improve the predicting power of the particular chosen model. While there are many advantages to this approach, there are a number of serious drawbacks. In particular, the probing of the data set depends on the particular model and somewhat less on the data. For problems where the model is over-parameterized, such as deep neural networks, the model is very expressive and can fit the data well (that is, interpolate it), the approach breaks (see \cite{KarzandNowak2019} and \cite{belkin2018reconciling} for a more thorough discussion). Furthermore, when using this approach to train a model, we use only the points that are probed and not the whole data set which may reduce the predictive power of the model, especially if the model requires a large number of data in order to be trained.

A different approach is to use a model-free method that aims to label the whole data set correctly by using  a few data points only, also known as semi-supervised learning \cite{ZhuSemi2008}, and label the whole data set by selecting some key data points to be labeled. After the data set has been labeled, a particular model can be trained on the {\bf complete} data set, and the errors (uncertainties) in the labeling process can be incorporated into the training routine. The advantage of this approach is that, as we see next, it leads to a very tractable problem that can be solved efficiently with some theoretical guaranties. It also allows for the use of the whole data set in the training process, which is advantageous especially for highly expressive models such as deep neural networks. Furthermore, the separation of the labeling and the training process allows the assessment of the data independent of the model, which enables training a number of models on the same data. Using an ensemble  of models can be important in many practical applications to obtain robust results and therefore, such an approach can be preferable to a model-dependent approach where the sampling is biased by the choice of a model.

Thus, in this work we treat the problem of active learning in two steps. In the first step we 
attempt to estimate the labels of the whole data set by probing it gradually and judiciously  and in the second step we train a model, in particular a deep network, using the complete pseudo-labeled data set. 
A somewhat similar approach that strongly influenced this work was proposed in \cite{Tur2005} and is refined in \cite{DasarathyNZ15} where a graph-based method was proposed in order to learn which data to label. \cite{DasarathyNZ15} labels points with the shortest distance between different labels. While our approach performs a pseudo-labeling of all samples and holistically chooses which point to label. This allows our approach to obtain theoretical bounds as well as practical assessment of the method, which minimizes the empirical risk, given a particular data set.

We note that the problem of "which data to annotate" in machine learning is very similar to the problem of "where to measure'' in the sensor placement problem and therefore we can use techniques that are developed for experimental design. In the sensor placement problem $\bfx$ is a physical location of a sensor (in 2 or 3D) and $\bfy$ is a function to be recovered, for example, temperature or pressure. Techniques for solving such problems are well understood and our work extends and uses these techniques in higher dimensions.

The rest of the paper is laid as follows. In Section~\ref{sec2} we  review  semi-supervised learning and develop an estimate to the errors associated with it under simple assumptions. In Section~\ref{sec3}  we use the errors in the estimated label and show how to reduce the estimated error by choosing the points to be labeled. This process is used in Optimal Experimental Design (OED) \cite{ChalonerVerdinelli1995,AtkinsonDonev1992,Pukelsheim93} and in particular OED of ill-posed problems \cite{HaberMagnantLuceroTenorio12,HaberHoreshTenorio09,HaberHoreshTenorio08}.
In Section~\ref{sec4} we  perform a number of numerical experiments and the paper is discussed and summarized in Section~\ref{sec5}.

\section{Semi-Supervised Learning by Graphical Methods}
\label{sec2}
Before we introduce our active learning approach, we first review how a small set of labeled data can be used in order to infer labels for the whole data set. 
The problem of labeling a large data set, given only a small number of labeled data is referred to as semi-supervised learning \cite{ZhuSemi2008}. Semi-supervised learning can be approached and solved in a number of different ways, but in the following we focus on graphical methods using a graph-Laplacian.


Given the data set $\bfX$ we assume that the true labels, $\bfy$, are unknown but we do have some observed labels, $\bfd^{obs}$, which are related to the true labels by
\begin{eqnarray}
\label{forwardy}
\bfd^{obs}_i = \bfy_i + \bfepsilon_i,
\end{eqnarray}
where $\bfepsilon_i$ is noise that we assume to be Gaussian but not necessarily uniform: 
\begin{equation}
\bfepsilon_i \sim N(0,\bfSigma),
\end{equation}
with $\bfSigma$ being the covariance of the noise.

Given that the data is noisy and only partially sampled, labeling the data is an ill-posed problem, and we therefore turn to regularization techniques to achieve this goal.
One way to regularize the problem is to use graphical methods. Let ${\boldsymbol \Delta}(\bfX)$ be the graph Laplacian of the data, that is:
\begin{align}
{\boldsymbol \Delta}_{ij} &= -\exp \left(-{\frac 1{\gamma}} \|\bfx_i - \bfx_j\|^2 \right), & {\boldsymbol \Delta}_{ii} = -\sum_{i \not= j} {\boldsymbol \Delta}_{ij},
\label{GL}
\end{align}
where $\gamma$ is a normalization parameter, we set to the median distance of all connections in the graph Laplacian.

The graph Laplacian can be used for regularization in a number of forms  \cite{von2007tutorial} \cite{nadler2009statistical, DSST2018} by introducing the quadratic form 
\begin{equation}
{\cal R}_m(\bfy) = \bfy^\top \bfL \bfy,
\label{eq:regularization_d}
\end{equation}
where
\begin{equation}
 \bfL =  ({\boldsymbol \Delta} + \tau \bfI)^{\eta}, 
 \label{eq:tau_eta}
\end{equation}
and the parameters $\tau$ and $\eta$ are hyper-parameters (see discussion in \cite{DSST2018}).

With the regularization defined, we can find approximate labels, $\widehat \bfy$, by solving the following weighted regularized least square 
\begin{eqnarray}
\label{optfory}
\widehat \bfy = {\rm arg}\min_{\bfy} \hf \|\bfW^{\hf}(\bfd -  \bfy)\|^2 + {\frac \alpha 2} \bfy^{\top}\bfL \bfy, 
\end{eqnarray}
where  $\alpha$ is a regularization parameter, $\bfW = {\rm diag}(\bfw)$ is a diagonal matrix with non-negative weights on its diagonal, and $\bfd$ is an extension of $\bfd^{obs}$ to all data points (the values in $\bfd$ outside $\bfd^{obs}$ does not matter since the weight of an unobserved data sample is zero).  
The first term in \eqref{optfory} is the (weighted) Mean Square Error (MSE) and the second term is a regularization term that biases the estimate toward a smooth solution as measured by the matrix $\bfL$.  
Note that we have used the MSE as loss function. Other loss functions such as cross entropy can be used as well however, they lead to non-trivial solutions that are more difficult to analyze (we discuss this in the next section). Thus, in this work we use the weighted MSE although extensions can be made to other loss functions. Using the MSE  it is easy to obtain the closed form solution
\begin{eqnarray}
\label{sol}
{\widehat \bfy} = (\bfW  + \alpha \bfL)^{-1} \bfW \bfd  = \bfH^{-1}\bfW \bfd,
\end{eqnarray}
where we have defined $\bfH = \bfW  + \alpha \bfL$.
The solution can also be thought of as the maximum a-posteriori estimate in a Bayesian inference, where the negative log prior is the regularization term (we discuss this point later), and the data covariance matrix is set to $\bfSigma^{-1} = \bfW$.

As previously noted, the problem is identical to the sensor placement problem \cite{HaberHoreshTenorio08}. In the sensor placement problem the data, $\bfx$, is typically in 2 or 3 dimensions (the physical location of the sensor), the unknown function (for example, temperature or pressure) is interpolated, given these measurements, the Laplacian is the discretization of the $\Delta = \partial_x^2 + \partial_y^2$ operator over the physical domain, and our goal is to estimate a smooth function, $\bfy(\bfx)$, over the physical domain. The techniques presented here are simple extension of the sensor placement problem into high dimensions where smoothness is measured by the graph Laplacian.

\bigskip

Let us now analyze the error in our estimate. The  solution, $\widehat \bfy$, depends on the data, $\bfd$, that has a stochastic noisy component and therefore, we can only estimate the expectation of the error.
We have that:
\begin{eqnarray}
 {\mathbb E}\, \|\bfy - \widehat \bfy\|^2 =  {\mathbb E}\, \|\bfy - \bfH^{-1} \bfW \bfd \|^2 = 
  {\mathbb E}\, \|(\bfI - \bfH^{-1} \bfW) \bfy - \bfH^{-1} \bfW \bfepsilon) \|^2.
\end{eqnarray}
When opening the norm and using the expected value we obtain the bias-variance decomposition \cite{tensiam}:
\begin{eqnarray}
 {\mathbb E}\, \|\bfy - \widehat \bfy\|^2 = \|(\bfI - \bfH^{-1} \bfW) \bfy\|^2 + 
 {\rm trace}( \bfW \bfH^{-2} \bfW \bfSigma)  = \| {\rm bias}\|^2 + {\rm var}.
 \end{eqnarray}
The bias can be simplified further: 
\begin{eqnarray}
 {\rm bias} = \bfy - \bfH^{-1} ( (\bfW  +\alpha \bfL) \bfy - \alpha\bfL \bfy) = \alpha\bfH^{-1} \bfL \bfy.
 \end{eqnarray}
From which we finally obtain:
 \begin{eqnarray}
 \label{exerr}
  {\mathbb E}\, \|\bfy - \widehat \bfy\|^2 =\alpha^2 \|\bfH^{-1} \bfL \bfy\|^2 +  {\rm trace}( \bfW\bfH^{-2} \bfW \bfSigma). 
  \end{eqnarray}
We note that the bias depends on the true solution (which is unavailable). However, the variance does not depend on the solution, but instead depends on the geometry of the data set, $\bfX$, alone.
In the next section, we shall see how equation \ref{exerr} can be recast and used in active learning.

\section{Optimal Experiment Design and Active Learning}
\label{sec3} 
 
The expected value of the errors in the estimated labels computed in \eqref{exerr}
depends on a number of quantities. First, the bias depends on the true values of the labels that we have no control over. Second, the errors depend on the matrix $\bfW$ which we control; if we set the $i$-th label to $\bfw_i=0$ then the $i$-th label is not sampled at all, and we treat this label as being unknown. 
Based on this observation, the active data sampling problem, is relaxed to the problem of setting weights to the data in order to obtain the highest accuracy in the estimate $\widehat \bfy$.
This problem of designing an experiment such that the expected error is minimized is referred  
to as A-optimal experimental design \cite{ChalonerVerdinelli1995,AtkinsonDonev1992,HaberHoreshTenorio08}, and in the context of active learning we refer to it as A-optimal active learning.

The solution to the problem without any constraint is trivial. If we are to estimate $\bfy$, then sampling every data $\bfx$ to obtain its label would give the most accurate result. This solution, however, ignores the cost of learning the labels of all data points. If, on the other hand, we recast the problem to include such a cost, by adding penalty on the one norm of $\bfw$, and for simplicity further assume that the data errors are iid with standard deviation $\sigma^2$ we obtain the following optimization problem
\begin{eqnarray}
\label{OEDr}  
\bfw^* = {\rm arg}\min \phi(\bfw)   &=& \alpha^2 \|\bfH(\bfw)^{-1} \bfL \bfy\|^2 + \sigma^2 {\rm trace}( {\rm diag}(\bfw)\bfH(\bfw)^{-2} {\rm diag}(\bfw)) + \beta \|\bfw\|_1  \nonumber \\
{\rm s.t.}\ \  && 0 \le \bfw .
\end{eqnarray}
\Eqref{OEDr} uses the 1-norm of $\bfw$ for regularization. As shown in \cite{HaberHoreshTenorio08}, this typically yields a sparse set of non-zeros in the vector $\bfw$, which implies few labels are needed.

\Eqref{OEDr} cannot be minimized as is because it depends on the unknown solution $\bfy$. We now discuss two different techniques, which make different assumptions on $\bfy$ that allow us to obtain an approximate solution to the problem.

\subsection{A-Bayesian Active Learning}

In Bayesian active learning, we assume to have a prior on $\bfy$. That is, we view $\bfy$ as a random variable with an associated Gaussian probability density of the form
\begin{equation}
\bfy \sim {\cal N}(0, \alpha\, \bfL^{-1}).
\end{equation}
We further assume that we can control the noise in the data and that the noise covariance matrix is $\bfSigma = \bfW^{-1}$.
Since $\bfy$ is a random variable, we can  eliminate it from \eqref{OEDr} by taking the expected value obtaining:
\begin{subequations}
\begin{align}
{\mathbb E}_{\bfy} \alpha^2 \|\bfH^{-1} \bfL \bfy\|^2 +  {\rm trace}( \bfH^{-2} {\rm diag}(\bfw)) &= \alpha\,  {\rm trace} (\bfH^{-2} \bfL) +{\rm trace}(\bfH^{-2} {\rm diag}(\bfw))  \\
 &= {\rm trace}(\bfH^{-1}).
\end{align}
\end{subequations}

Thus, the A-Bayesian active learning or A-optimal Bayesian experimental design problem is reduced to a simple convex optimization problem of the form
\begin{eqnarray}
\label{OEDb}  
\min_{\bfw}\ \  &&{\rm trace}(\bfH(\bfw)^{-1} )  + \beta \|\bfw\|_1 \\
\nonumber
{\rm s.t.}\ \  && 0 \le \bfw \le 1 .
\end{eqnarray}

Bayesian designs are typically a "one shot" active learning since the design depends on the geometry of the data $\bfX$ and not on its labels. 
While the final expression in \eqref{OEDb} is very "clean" mathematically, it has many practical short-comings. First, the assumption that we can control the noise, that is, assuming that the data covariance matrix is $\bfW^{-1}$, is unrealistic in many cases. Second, the notion  that $\bfy$ is Gaussian with an inverse covariance matrix $\alpha \bfL$ is very strong and much more informative than the assumption that $\bfy$ is smooth \cite{tensiam,st}. 
As we see next, there are other ways to deal with the bias that are less restrictive, and lead to different designs that do depend on the observed labels. Nonetheless, when exploring a data set for the first time, before probing any labels, Bayesian design can be useful in giving a first guess to what data should be initially labeled.

\subsection{Adaptive Active Learning and  Frequentist A-Optimal Design}

The main difference between the classic Optimal Experimental Design and the Active Learning problem is the ability to further probe the data and ask for new labels at a later time. This can make the Active Learning process better as we are able to iteratively estimate the bias term. Thus,
 a different approach for the solution of the problem is to recall that in practice, we are not interested in a single $\bfw$ but rather solve for it gradually and iteratively, that is, we typically start by using a small cost $C$ to obtain only a few labels (this can be done using the Bayesian design). Next, we use these labels to estimate all the labels by solving the optimization problem \eqref{sol} and then, decide where should we sample next. The process repeats iteratively until the data is labeled to sufficient accuracy as determined by the user, or until the budget for labeling is diminished.    
Given this gradual process, we actually have an estimate for $\bfy$ at every step of the 
computation. This estimate can be used in \eqref{OEDr} in order to estimate the bias term. Using the  knowledge we obtained on the labels by using coarse sampling, allows us to find a design that is adapted to the particular labels at hand. A similar idea for the estimation of the bias is also presented in \cite{Stark08,Stark92}.
The process can be thought of as a gradual refinement method, where we decide where to refine based on the solution obtained on a coarser discretization of the data set. The gradual refinement algorithm that results from the process discussed above is presented in 
Algorithm~\ref{alg1}.

The bulk of the computation in this algorithm (besides the true labeling) is solving the design problem \eqref{OEDr}
and in estimating the solution \eqref{sol}. Efficient numerical solutions to these problems is discussed in \cite{HaberMagnantLuceroTenorio12,HaberHoreshTenorio08}. 

\begin{algorithm}{$\widehat \bfy, \bfw ={\rm AdpativeActiveLearning }(\bfX)$}
\begin{algorithmic}
\State Initialize: $\widehat \bfy=0$
\State Compute the graph Laplacian and the regularization matrix $\bfL(\bfX)$
\While{not converged}
\State Set $\bfy = \widehat \bfy$ and solve \eqref{OEDr} for $\bfw^*$ 
\State Probe the data set based on $\bfw^*$ 
\State Estimate $\widehat \bfy(\bfw^*)$ by solving \eqref{sol}
\State Assess convergence
\EndWhile
\end{algorithmic}
\caption{Adaptive Active Learning. \label{alg1}}
\end{algorithm}

Before we present some numerical experiments we comment on a number of important points.

\begin{itemize}
\item {\bf Approximately solving  \eqref{OEDb} and \eqref{OEDr} } \\
Both problems require an estimation of the trace of large matrices. In this work we use stochastic trace estimators of the form
\begin{eqnarray}
\label{ste}
{\rm trace} (\bfA^{-1}) \approx {\frac 1N} \sum \bfv_j^{\top} (\bfA^{-1} \bfv_j)
\end{eqnarray}
where $\bfv_j$ is a random vector of $\pm 1$.

Note that the matrix $\bfA^{-1}$ need not be formed or stored. Only matrix vector products of the form $\bfA^{-1}\bfv_j$ need to be computed, and this can be done by solving linear systems of the form:
\begin{equation}
\bfA \bfz_j = \bfv_j. 
\end{equation}
Since in our application $\bfA$ is sparse, sparse linear algebra techniques are used. For moderate to large problems we use Cholesky factorization and for very large problems preconditioned conjugate gradient method is used to solve the system. Using stochastic trace estimators with sparse linear algebra allows for an efficient computation of the objective functions to be used in all optimization routines.
\item {\bf Common Extensions} \\
In the process above we have used the weighted MSE however, in many cases one uses the cross entropy as a loss function. In this case the estimate $\widehat \bfy$ cannot be obtained analytically and there is no closed form expression to the bias and variance. Nonetheless, it has been proposed in \cite{HaberHoreshTenorio08,ChalonerVerdinelli1995} to either use local estimates of the Hessian or a Monte-Carlo simulation to estimate the error. Both lead to similar formulations as \eqref{OEDr} but with less theoretical guarantees. 

\item {\bf Training a model with estimated labels} \\
Assume now that we would like to use the labeled obtained by \eqref{sol} in order to train a model
\begin{eqnarray}
\label{model}
\widehat\bfy  \approx f(\bfx,\bftheta),
\end{eqnarray}
where $f$ is a model that depends on the trainable parameters $\bftheta$.
In this case, the estimated labels contain correlated errors and this needs to be taken under consideration. It is straight forward to see that the covariance of the estimated errors is nothing but
$\bfH(\bfw)^{-1}$. While it is difficult to work with this covariance matrix, it is possible to approximate its diagonal in order to use different variances for the data, making sure that data that has large uncertainty is not fit to the same degree of data with large certainty.

\item {\bf Assigning multiple labels each iteration} \\
\label{multisamples}
Using Algorithm \ref{alg1} to select one new data sample for labeling each iteration is simple, take the sample with the largest value in $\bfw^*$. 
However, since $\bfw^*$ tends to be smooth over the dataset the largest values in $\bfw^*$ tend to be neighbours, as such additional steps are generally needed in order to get multiple sensible data samples for labeling each iteration. Our pragmatic solution to this problem is that we take the one with largest value in $\bfw^*$, and once a sample is selected for labeling, we exclude all neighbouring samples from the selection process during this iteration.
\end{itemize}

\section{Numerical Experiments}
\label{sec4}
 
We present two different examples: A simple 2D example, which gives intuitive understanding, and a classical benchmark, MNIST. 
In both examples we only used the nearest 10 neighbours when computing the graph Laplacian.

\subsection{A simple 2D example}
\begin{figure}[htb!]
	\begin{center}
    \begin{subfigure}[htb!]{0.3\textwidth}
      \includegraphics[width=\textwidth]{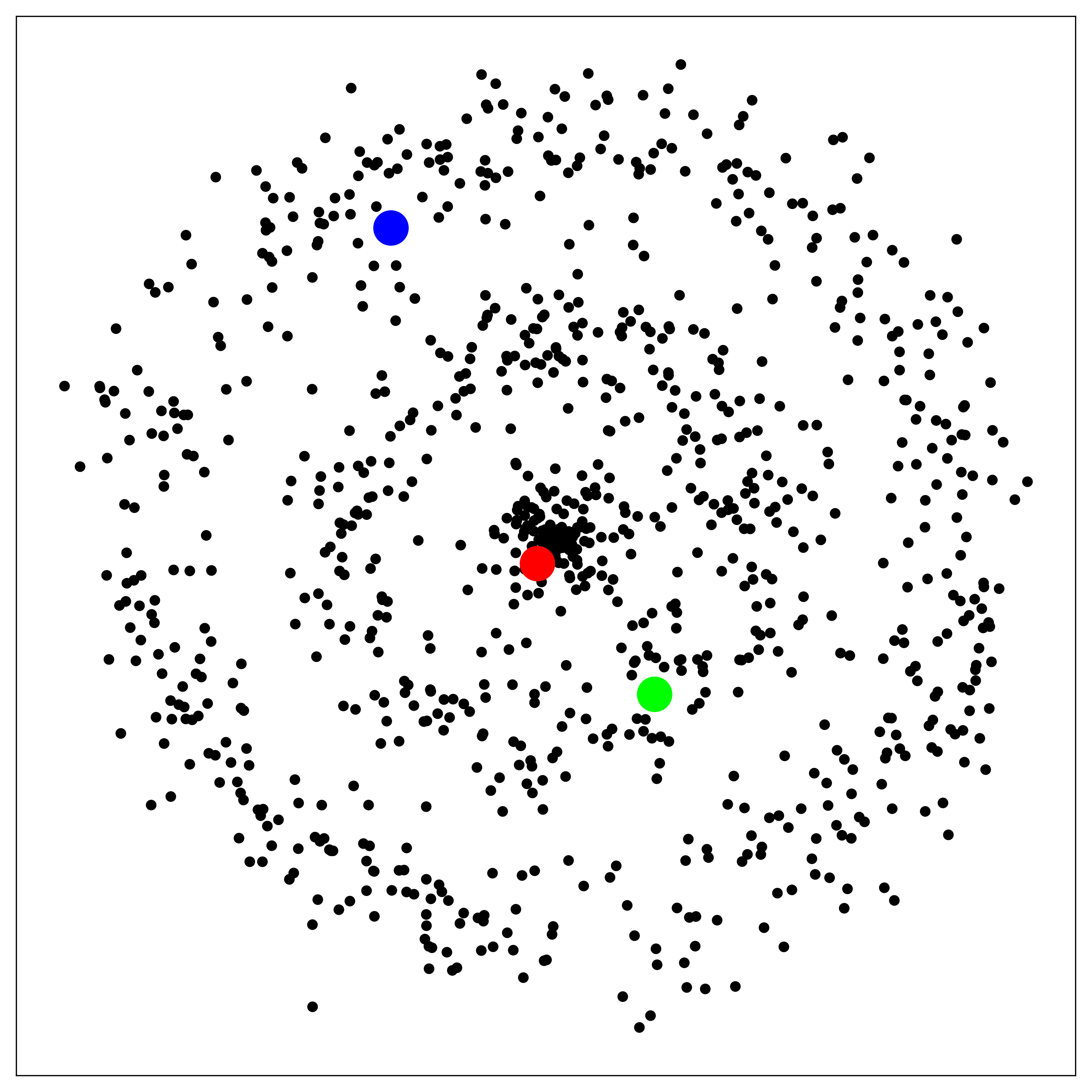}
      \caption{Input data}
      \label{fig:ex1_input}
    \end{subfigure}
    \begin{subfigure}[htb!]{0.30\textwidth}
      \includegraphics[width=\textwidth]{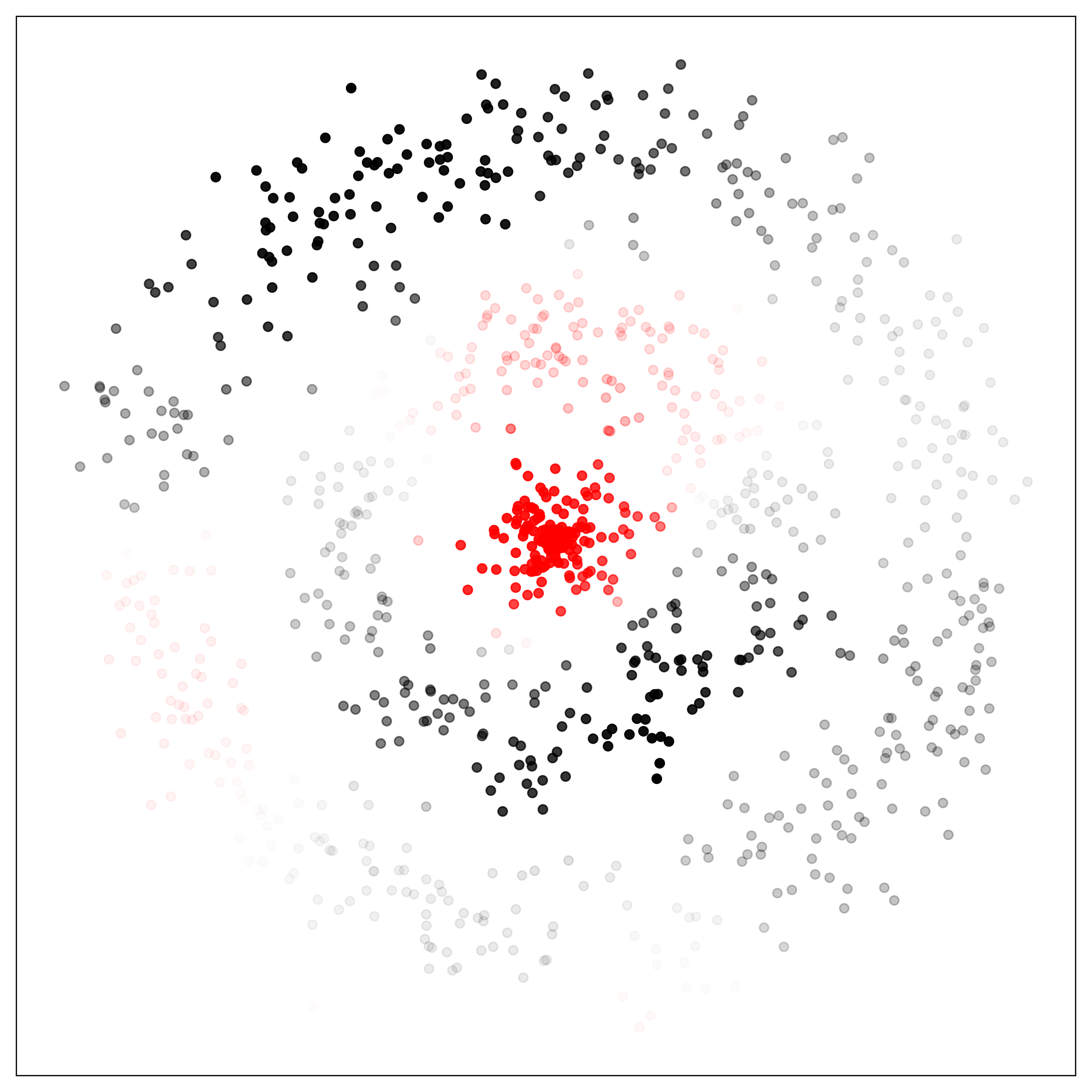}
      \caption{1 vs all labels}
      \label{fig:ex1_1vsall}
    \end{subfigure}
    \begin{subfigure}[htb!]{0.30\textwidth}
      \includegraphics[width=\textwidth]{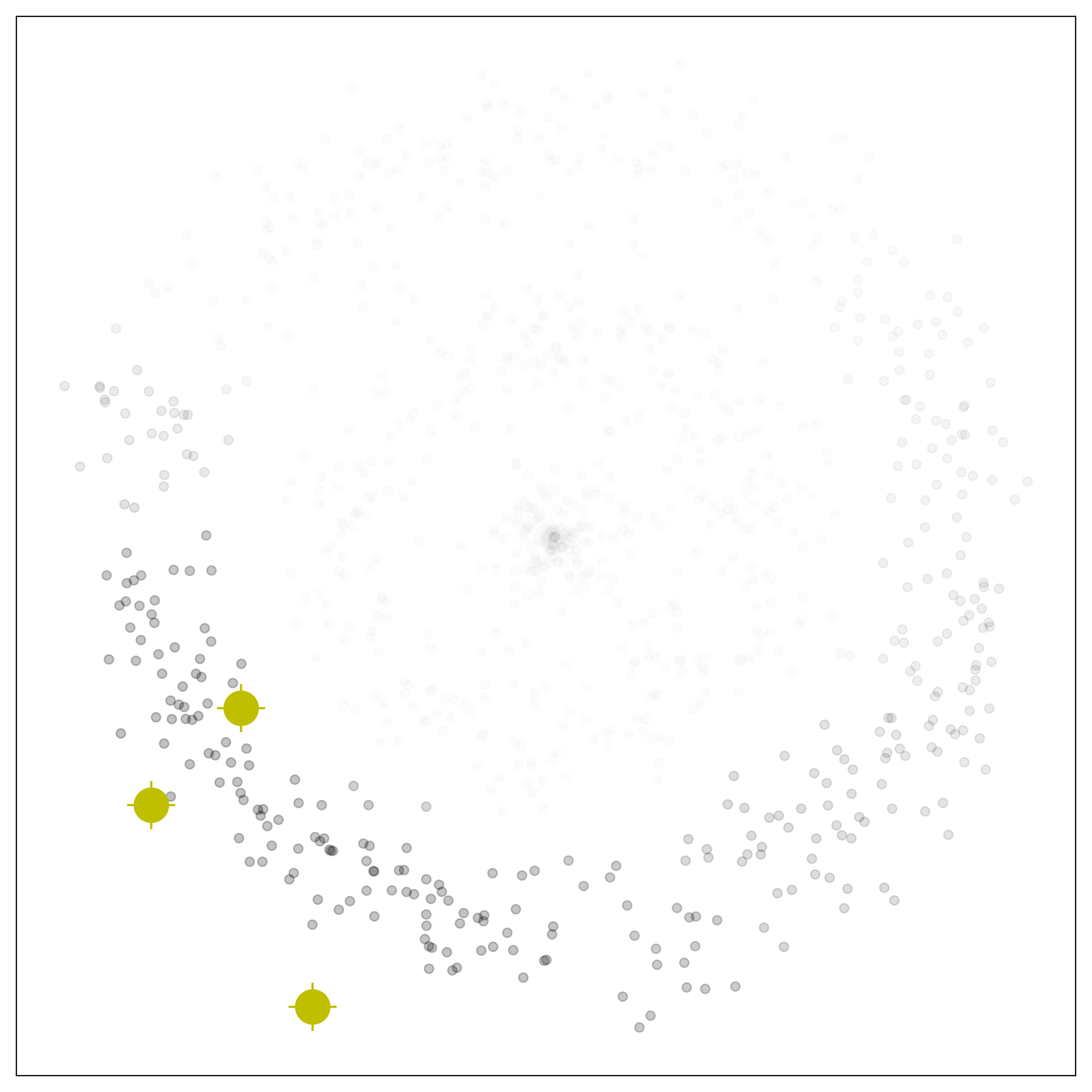}
      \caption{$|\frac{\partial \phi}{\partial \bfw}|$}
      \label{fig:ex1_df}
    \end{subfigure}
    \end{center}
\caption{Shows the initial states. \ref{fig:ex1_input} shows the input data, with the 3 labeled data points represented by enlarged colored dots. \ref{fig:ex1_1vsall} shows the label predictions made for class 1 versus [2,3]. The active class is shown as red points, while other classes are grouped together as black points. The opacity of the points shows their label certainty. \ref{fig:ex1_df} visualizes the derivative of the objective function, based on the labels predicted in \ref{fig:ex1_1vsall}, with the opacity showing the magnitude of each individual points contribution. The yellow dots indicate the points selected for labeling.}
\label{fig:ex1_init}
\end{figure}

\begin{figure}[htb!]
	\begin{center}
    \begin{subfigure}[htb!]{0.3\textwidth}
      \includegraphics[width=\textwidth]{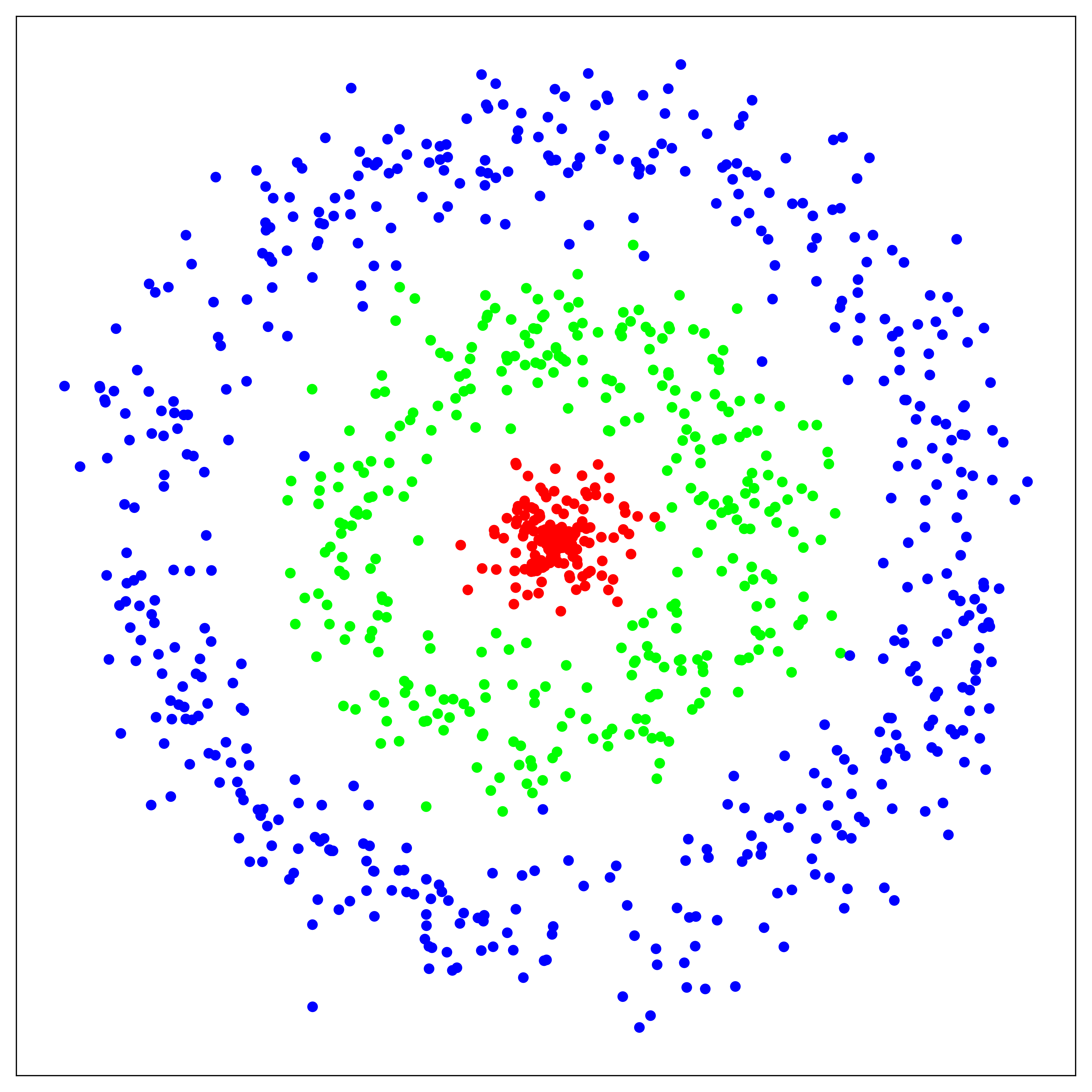}
      \caption{True labels}
      \label{fig:ex1_true}
    \end{subfigure}
    \begin{subfigure}[htb!]{0.30\textwidth}
      \includegraphics[width=\textwidth]{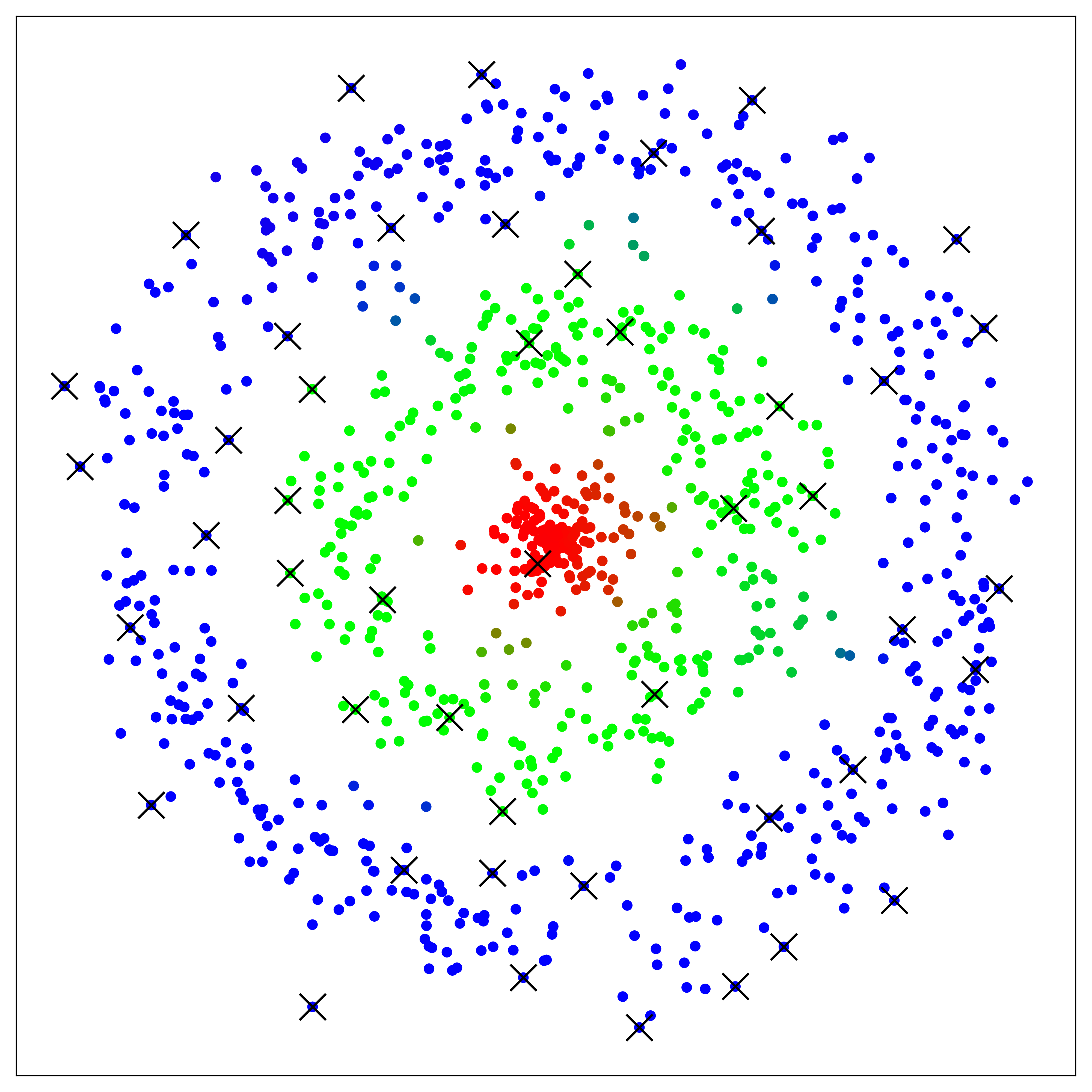}
      \caption{Predicted labels}
      \label{fig:ex1_pred}
    \end{subfigure}
    \begin{subfigure}[htb!]{0.30\textwidth}
      \includegraphics[width=\textwidth]{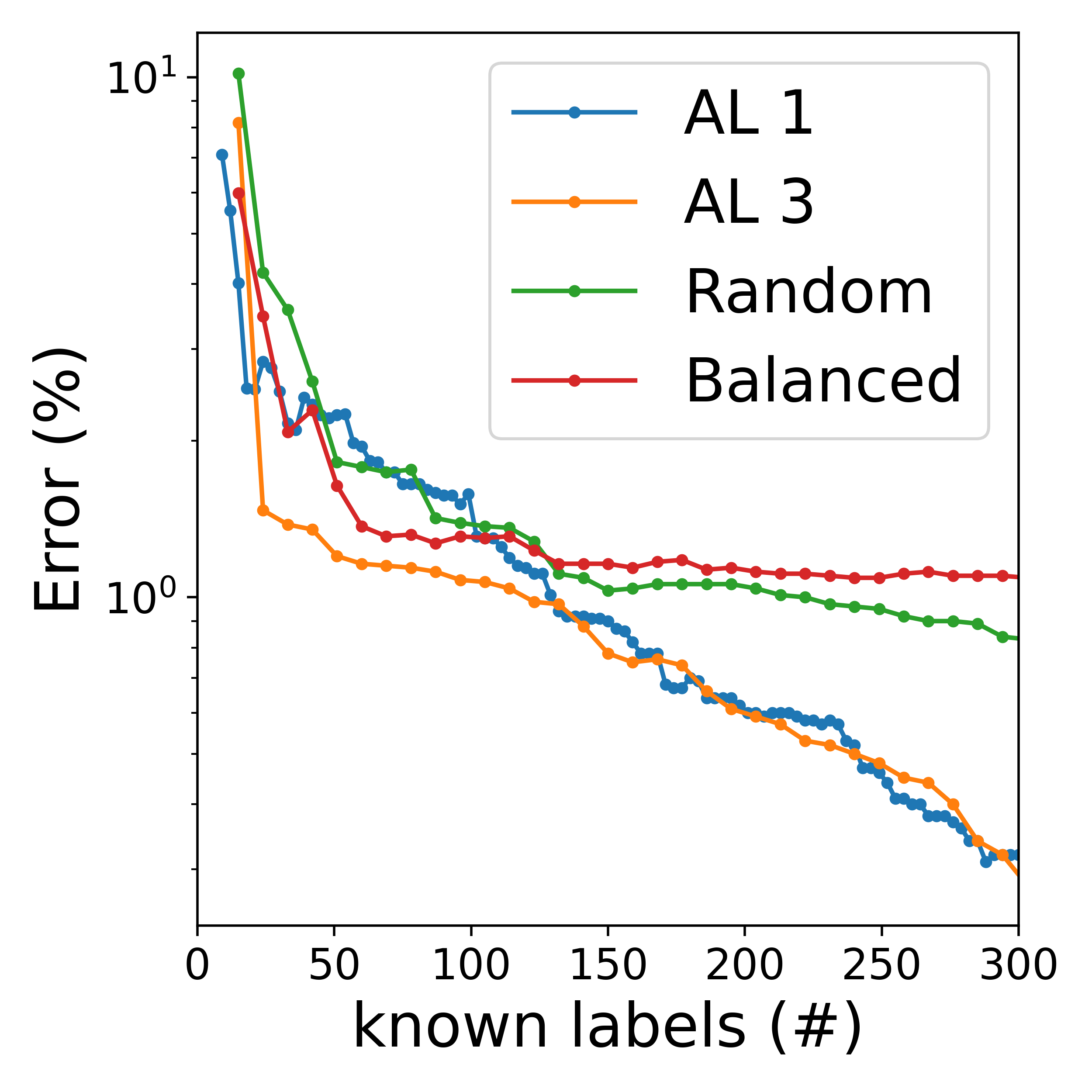}
      \caption{Method comparison}
      \label{fig:ex1_results}
    \end{subfigure}
    \end{center}
\label{fig:ex1_outputs}
\caption{Outputs from the numerical experiments with a simple data set. \ref{fig:ex1_pred} shows the predicted labels, with each $\times$ indicating a labeled data point. \ref{fig:ex1_results} shows a comparison of different methods averaged over 10 runs. "AL 1" is our active learning approach given in algorithm~\ref{alg1}, with 1 point selected at each iteration, while "AL 3" is our method with 3 points selected at each iteration. "Random" is randomly selected labeling points. "Balanced" also randomly selects points each iteration, but balanced with an equal number of labeled data points from each class.}
\end{figure}

We are given a data set with 1000 data points, three classes, and 2D features, as visualized in Figure~\ref{fig:ex1_input}, with corresponding true labels shown in Figure~\ref{fig:ex1_true}.  

The data set is initialized with one known labeled point from each class. We adaptively learn new labeled points and cluster the data, as outlined in algorithm~\ref{alg1}.

For the first iteration and class, Figure~\ref{fig:ex1_1vsall} visualizes the predicted labels, while Figure~\ref{fig:ex1_df} visualizes each individual point's absolute contribution to the objective function's derivative, $|\frac{\partial \phi}{\partial \bfw}|$, from which the next labeled points are chosen (in this case 3 points are chosen each time).  
Figure~\ref{fig:ex1_pred} shows the data points selected for labeling, and labels predicted, at a later iteration of the algorithm. 
We run the algorithm until 300 labels have been selected. Figure~\ref{fig:ex1_results} shows the error of our method compared to other selection schemes. 
It is clear that we are able to train any model on the recovered data set.

\subsection{MNIST}
\label{MNIST}
We next test our algorithm on the MNIST data set, where we use the standard segmentation of the data into 60,000 training images and 10,000 test images. For the initial graph Laplacian generation, we transform the image features with a fully-connected autoencoder to a latent 50-dimensional space. 
We start with 2 randomly known labels from each class and follow algorithm~\ref{alg1}. 

\begin{figure}[htb!]
	\begin{center}
    \begin{subfigure}[htb!]{0.3\textwidth}
      \includegraphics[width=\textwidth]{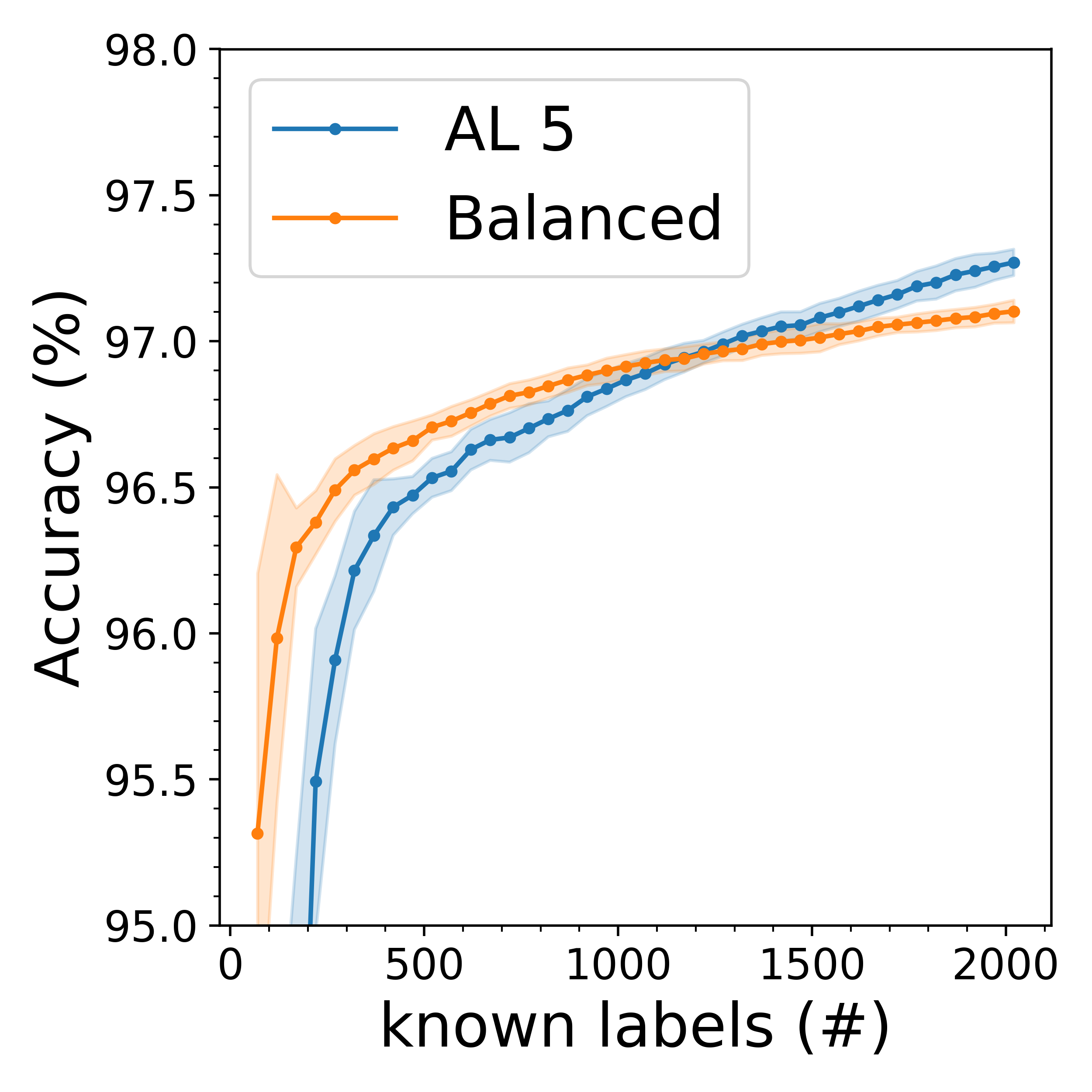}
      \caption{Method comparison}
      \label{fig:mnist_1}
    \end{subfigure}
    \begin{subfigure}[htb!]{0.3\textwidth}
      \includegraphics[width=\textwidth]{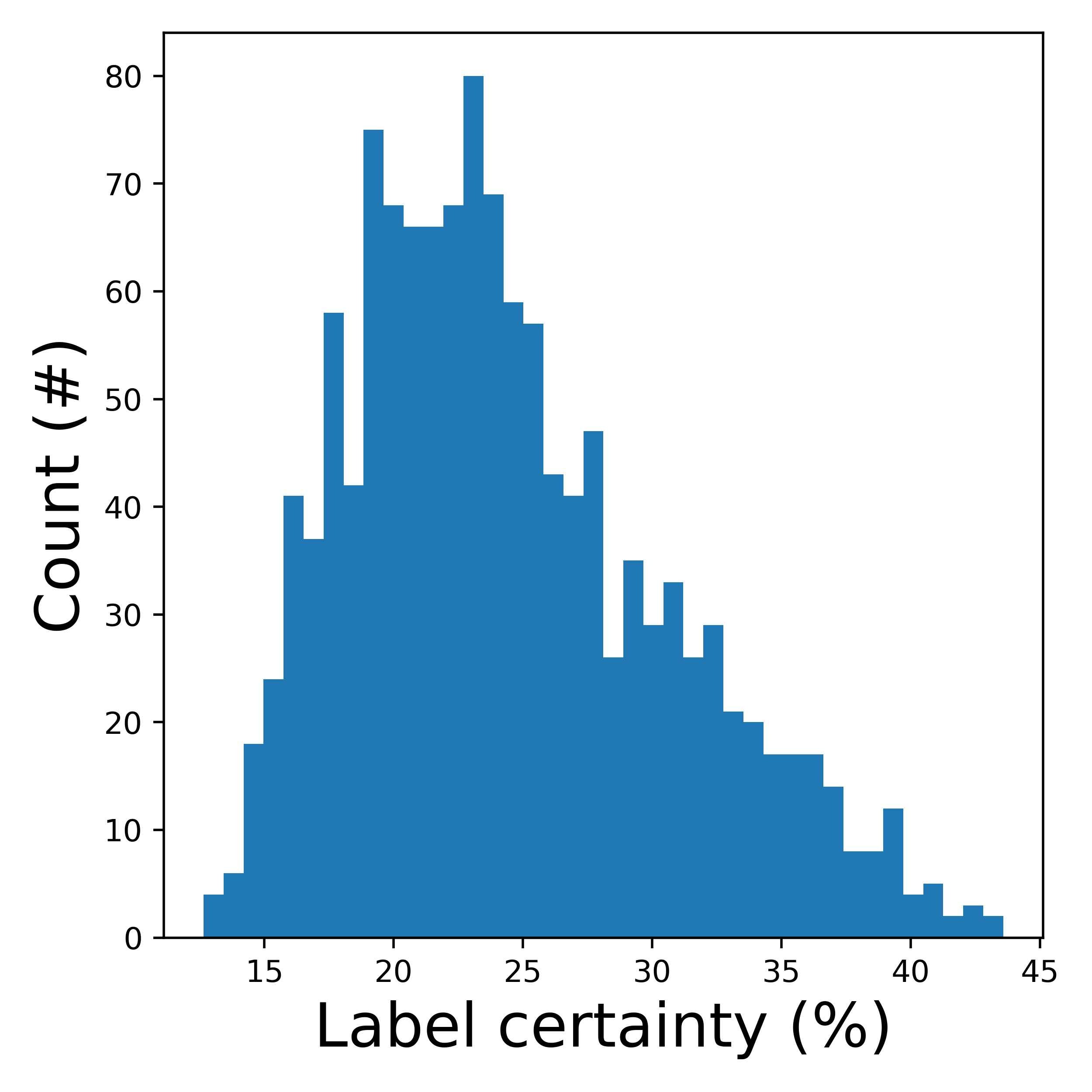}
      \caption{Wrong labels}
      \label{fig:mnist_2}
    \end{subfigure}
    \begin{subfigure}[htb!]{0.3\textwidth}
      \includegraphics[width=\textwidth]{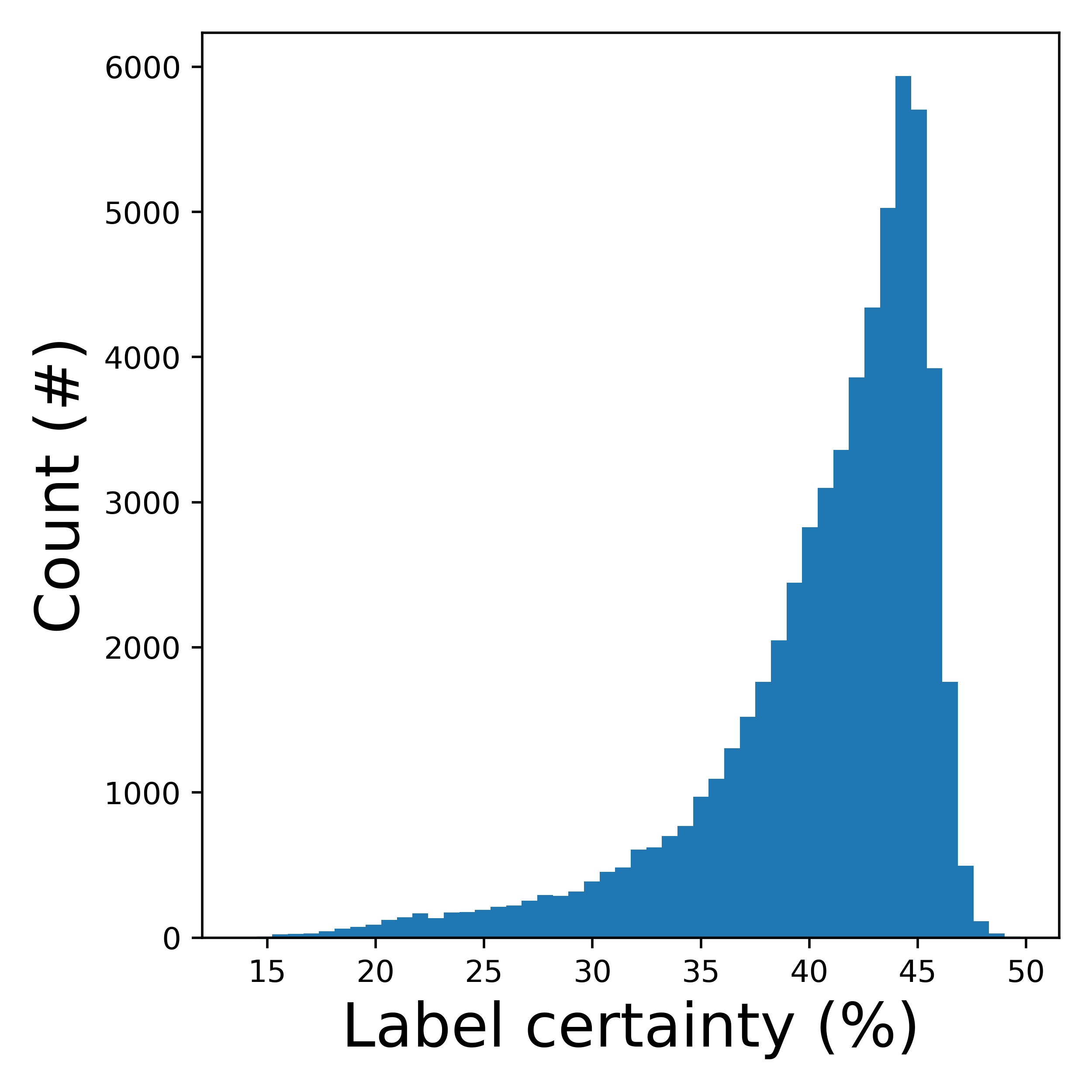}
      \caption{Right labels}
      \label{fig:mnist_3}
    \end{subfigure}
    \end{center}
\caption{Results for MNIST. \ref{fig:mnist_1} shows the clustering accuracy on the training set. "AL 5" is adaptive active learning with 5 labels learned at each iteration.  "Balanced" is random labeling with an equal number of labeled data points from each class. The lines represent the mean accuracy after 8 runs, while the coloured regions represent the standard deviation. \ref{fig:mnist_2}/\ref{fig:mnist_3} shows the label certainty predicted by the neural network for all the wrongly/correctly classified labels, at the end of a single adaptive active learning run.}
\label{fig_mnist}
\end{figure}

Once we have labeled 2020 points, we use ResNet18, as described in \cite{he2016deep}, to learn the transduced clustering labels. We allow the network to train for 20 epochs, and use a cosine ramped learning rate \cite{loshchilov2016sgdr} to ensure that the network ends with sensible weights. As optimizer we use stochastic gradient descend with momentum set to 0.9 and a weight decay of $10^{-5}$. Afterwards we test the trained neural net on the 10000 test samples, from which we get the following accuracies: "AL5"$= 97.68 \pm 0.18 \%$, "Balanced"$= 97.63 \pm 0.07 \%$. 
Each approach was rerun 10 times, to produce the error bounds shown here.

Figure~\ref{fig:mnist_1} compares our adaptive active learning approach to random and balanced sampling, despite the much higher dimensionality of this data set compared to the simple 2D example shown previously, we once again see a  superior learning rate at late stages.
Figure~\ref{fig:mnist_2}-\ref{fig:mnist_3} shows the label certainties for correctly and wrongly classified images produced by the neural network at the end one of our adaptive active learning runs, which are essentially identical to the cluster-predicted label probabilities. 
We see that the confidence in the correctly classified labels  is overall much higher than the confidence in the wrongly classified labels. This is important for the predictive power of our method, and for the methods ability to pick relevant data points to label. 

\subsection{Comparison to other Active learning approaches}
We compare our approach to the pseudo-labelling approach suggested by \cite{wang2016cost}, which ranks unlabeled samples in an ascending order according to the first and second most probable class labels as predicted by the classifier, and selects the most uncertain ones for labelling. The psuedo-labelling is done by a process called high-confidence psuedo-labelling, which assigns a label to those datapoints for which the classifiers labeling entropy is less than a threshold value $\delta$. 

Since \cite{wang2016cost} uses cross-entropy loss, we employ cross-entropy loss for all methods during this test. Once again we test on the MNIST data set, with settings similar to the ones in section \ref{MNIST}, except as mentioned below. We initiate with 10 randomly known labels from each class, and we use a static learning rate of 0.005 with the ADAM optimizer. For the initial training of the classifier we allow 50 epochs, but afterwards we train for 20 epochs after each set of new labels are added, which happens 50 times. For the active learning approach from \cite{wang2016cost}, we set the threshold hyperparameter $\delta = 0.005 - 8 \cdot 10^{-5} t$, where $t$ is the current iteration. This ensures that the amount of samples being pseudo-labelled are in accordance with the rates reported in \cite{wang2016cost}.

\begin{figure}[htb!]
	\begin{center}
    \begin{subfigure}[htb!]{0.4\textwidth}
      \includegraphics[width=\textwidth]{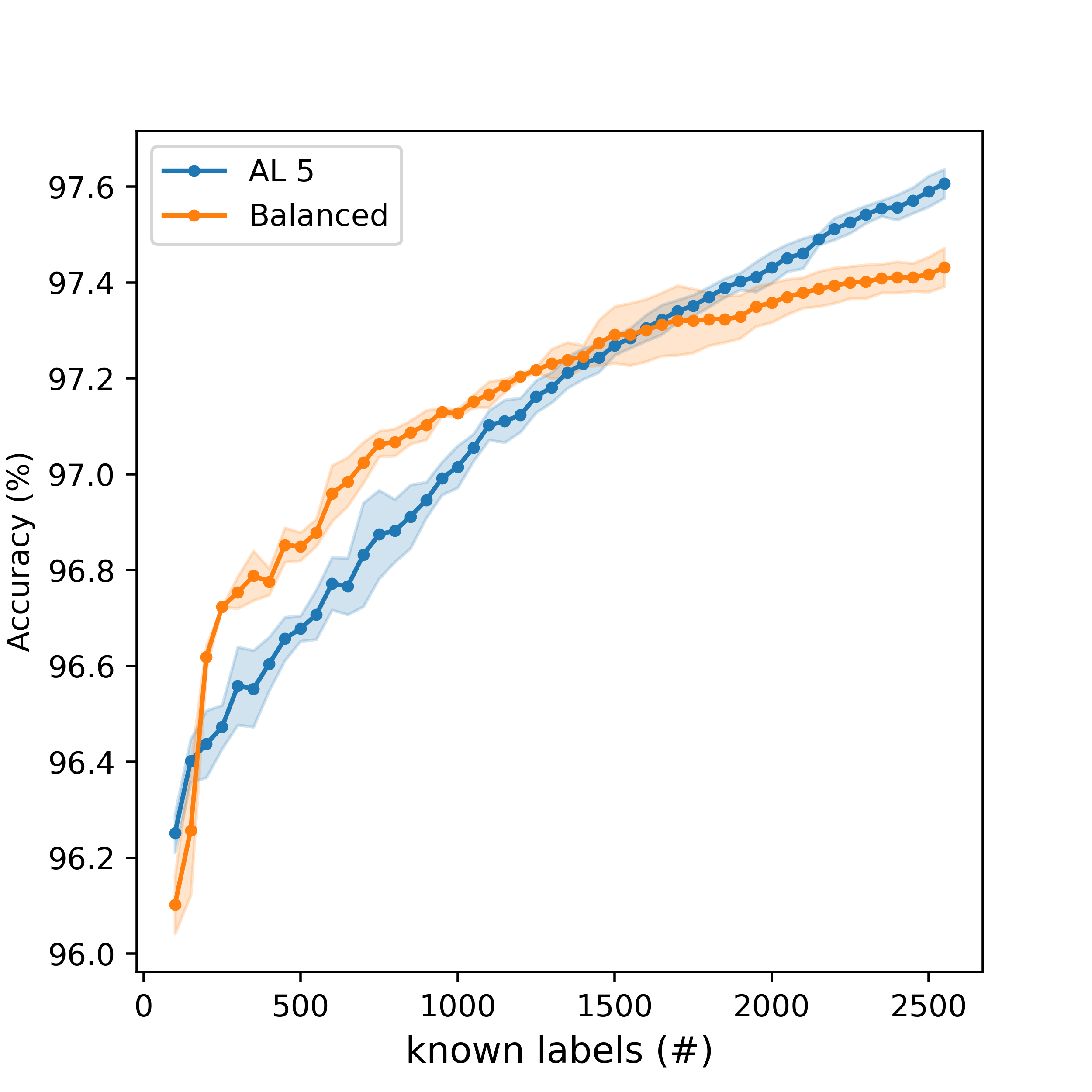}
      \caption{Clustering accuracy}
      \label{fig:mnistcomp1}
    \end{subfigure}
    \begin{subfigure}[htb!]{0.4\textwidth}
      \includegraphics[width=\textwidth]{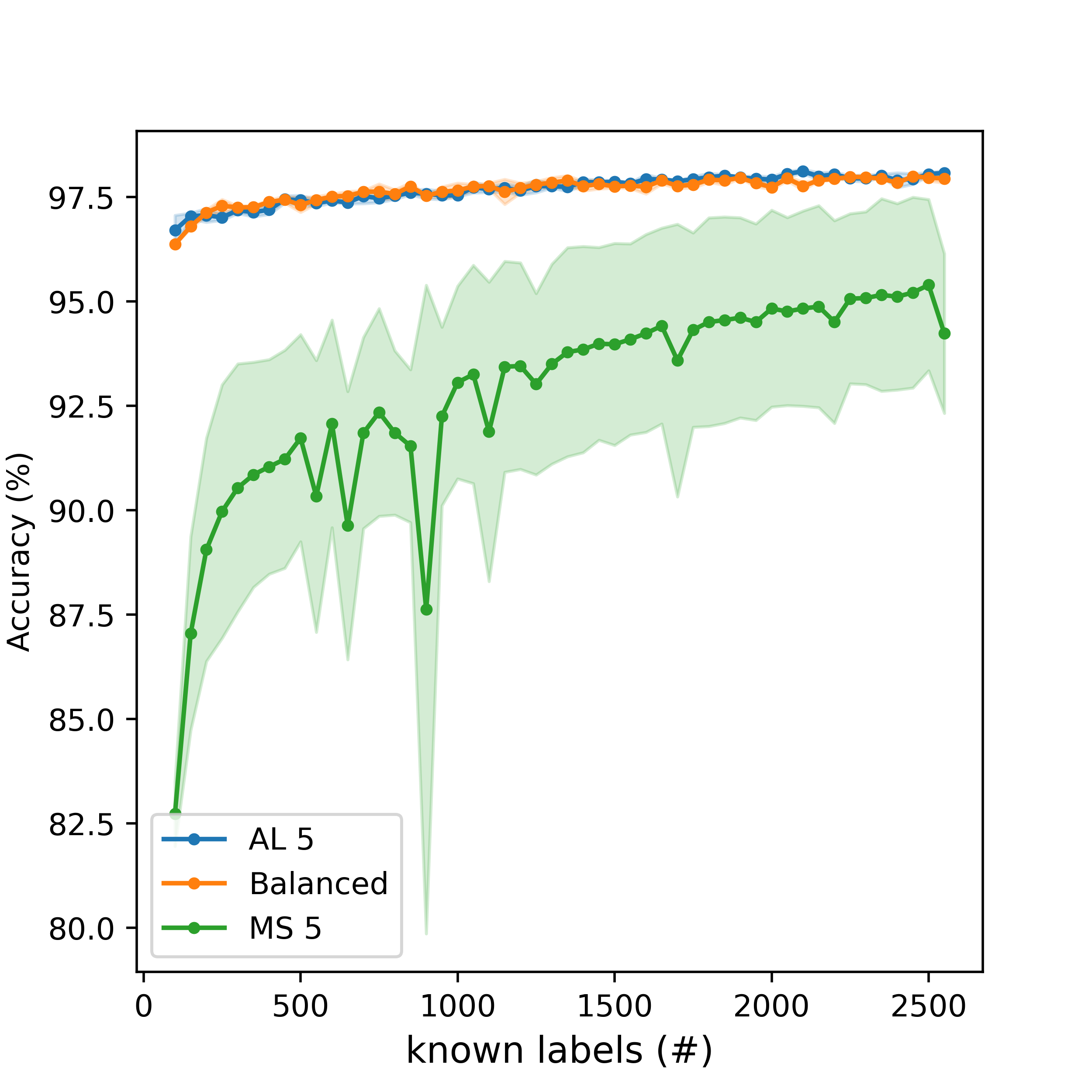}
      \caption{Test accuracy}
      \label{fig:mnistcomp2}
    \end{subfigure}
    \end{center}
\caption{Results for MNIST. \ref{fig:mnistcomp1} shows the clustering accuracy on the training set, while \ref{fig:mnistcomp2} shows the test accuracy for the three different methods. The shaded area represents one standard deviation, and the results are based on three repetition.}
\label{fig_mnist2}
\end{figure}

The results from this comparison can be seen in Figure \ref{fig_mnist2}. Figure \ref{fig:mnistcomp1} shows the clustering accuracy which shows a similar picture to Figure \ref{fig:mnist_1}, while \ref{fig:mnistcomp2} shows that our approach achieves a much higher test accuracy compared to the approach suggested in \cite{wang2016cost}. 

\section{Discussion and Summary}
\label{sec5}

In this work we have presented a methodology adopted from A-optimal experimental design for the field of active learning. Our approach  is based on two distinct stages. In the first stage we estimate the labels and in the second, we use the estimated labels to train a model. The decision, where to sample, is based on reducing the estimated error in the recovered labels. To this end we use the graph Laplacian for regularization and estimate the recovery error based on this regularization. We then use the estimated errors and approximately solve a sequence of  optimization problems that aim to iteratively reduce the errors.

We have validated our results on a simple 2D data set and on the MNIST data set and compared it to another well-known active learning scheme employing pseudo-labelling. 
Based on figure~\ref{fig:mnistcomp2} our semi-supervised clustering pseudo-labelling scheme seems superior to the high-confidence pseudo-labelling scheme employed by \cite{wang2016cost}. However, given that the passive learning shows a similar test accuracy we recognize that this is more a case of our semi-supervised pseudo-labelling rather than the active learning. 
A more direct comparison with other active learning approaches would be desireable, but due to the ingrained nature of our scheme, the active learning component is only guaranteed to be optimal if used in conjunction with the semi-supervised pseudo-labelling. 

While active learning can yield superior results compared with simple random sampling, our comparisons shown in figure~\ref{fig:ex1_results} and figure~\ref{fig:mnist_1}, demonstrate that for our approach this occur only in later stages of the learning.  
This is also seen in other active learning approaches \cite{KarzandNowak2019}.
This phenomena is also known from similar problems in optimal design and in mesh refinement when a function is sampled iteratively to obtain higher accuracy. The main reason is that our initial assumption is based on the recovery using the graph Laplacian. If the true solution is not sufficiently smooth as measured by the graph Laplacian then, initial sampling that use the assumption heavily will yield biased results. Choosing better regularization, or using the graph Laplacian on features of the data, such that it yields a better graphical description of the data is an open research question.



\section{Data availability statement}
Our code is available at \url{https://github.com/tueboesen/A-Optimal-Active-Learning}
\printbibliography

\appendix
\section{Hyperparameters}
The method presented above contains a variety of hyperparameters which are not essential to the understanding of the method, but still needs to be set correctly in order for the method to work. The values used in our numerical experiments can be seen in Table~\ref{tab:hyperparam}.

Equation \ref{eq:tau_eta} contains the hyperparameters $\tau$ and $\eta$. $\tau$ ensures that the Laplacian is semi-positive definite, which is needed for stability, while $\eta$ gives the order of the Laplacian. $\tau$ should be small in order to not skew the Laplacian significantly but positive and large enough to ensure stability. $\eta$ should be a small positive integer (1-3), if $\eta$ becomes too large then the computation of the Laplacian becomes too expensive.

Equation \ref{optfory} introduces $\alpha$ which is a regularization hyperparameter that balances the smoothness of the predicted labels. Generally, this parameter is left as unity.

Equation \ref{OEDr} introduces the parameters $\sigma$ and $\beta$. Where $\sigma^2$ is the standard deviation of the data errors tuning the noise. $\sigma$ should be chosen according to the data. In our examples we assumed a relatively small noise with $\sigma=10^{-2}$. $\beta$ tunes the cost of labelling additional samples. In the adaptive approach which we primarily use in this paper we find that tuning $\beta$ does not matter since we are iteratively selecting a fixed number of new samples anyway. 
\begin{table}[h!]
\begin{center}
\begin{tabular}{ |rr| } 
 \hline
 $\tau=$ & $10^{-2}$\\ 
 $\eta=$ & $2$ \\
 $\alpha=$ & $1$ \\
 $\sigma=$ & $10^{-2}$ \\
 $\beta=$ & $0$ \\
 \hline
\end{tabular}
\caption{The hyperparameter values used in our numerical experiments.}
\label{tab:hyperparam}
\end{center}
\end{table}

\end{document}